\documentclass[conference]{IEEEtran}
\IEEEoverridecommandlockouts
\usepackage{cite}
\usepackage{amsmath,amssymb,amsfonts}
\usepackage{algorithmic}
\usepackage{graphicx}
\usepackage{textcomp}
\usepackage{xcolor}
\def\BibTeX{{\rm B\kern-.05em{\sc i\kern-.025em b}\kern-.08em
    T\kern-.1667em\lower.7ex\hbox{E}\kern-.125emX}}
\begin{document}

\title{Detection of healthy and diseased crops in drone captured images using Deep Learning
}

\author{\IEEEauthorblockN{ Jai Vardhan }

\textit{IIIT Naya Raipur}\\
jai20100@iiitnr.edu.in \\
\IEEEauthorblockA{\textit{Computer Science and Engineering} 
}
\and
\IEEEauthorblockN{Kothapalli Sai Swetha}

\textit{IIIT Naya Raipur}\\
kothapalli20102@iiitnr.edu.in\\
\IEEEauthorblockA{\textit{Electronics and Communication Engineering} }
}

\maketitle

\begin{abstract}
Plant health and food safety are inextricably linked. Everyone is worried about the condition of green plants. Plant diseases disrupt or modify a plant's important activities by interfering with its normal state. The proposed approach aids in the detection of plant diseases.The database gathered from the Internet is appropriately separated, and the various plant species are recognized to obtain a test database containing numerous plant diseases that are used to analyze the project's correctness and confidence level.Then, using training data, we will train our classifier, and the output will be predicted with the highest possible accuracy. We employ a Convolution Neural Network (CNN), which is made up of several layers that are used for prediction. CNN's outperform other technologies in problems involving crop disease categorization and detection. They can handle complex challenges under harsh imaging conditions.
A prototype drone model is being utilized for live monitoring of huge agricultural fields, with a high-resolution camera attached to record photographs of the plants, which will be used as input to determine whether the plant is healthy or not.
\end{abstract}


\section{Introduction}

Agriculture is a vital industry in India, as it is the main source of livelihood for a majority of the population. It is a major contributor to the country's economy, accounting for approximately 16\% of India's GDP and employing around 50\% of the country's workforce. Agriculture also plays a crucial role in meeting the food and nutritional needs of the growing population. Additionally, the export of agricultural products is an important source of foreign exchange for India. In short, the importance of agriculture in India cannot be overstated. Nowadays, diseases in plant leaves are a major problem that have a negative impact on plant longevity and the output of high-quality crops. Additionally, using just your eyes to determine the leaf's current state is exceedingly challenging. This lowers the production of crops of good quality. We intended to apply a deep learning-based strategy to segment, to pick out each little portion of the leaf, and to detect the illness, as well as to evaluate the quality of the plant, in order to solve this issue. The primary operations are data collection, processing, and variable rate of input application. Plant health and food safety are inextricably linked. Plant health is a concept that is commonly used yet poorly defined. Image processing can be thought of as a subset of signal processing. The input in image processing is in the form of an image, such as a photograph or video format. Image processing will produce either an image or a set of features or metrics relevant to the provided image.
The traditional method for detecting and recognising plant diseases is based on naked-eye inspection, which is a sluggish and inaccurate method. Due to the scarcity of expertise in some countries, contacting experts to determine plant disease is costly and time-consuming. Irregular plant inspection results in the development of many illnesses on the plant, which necessitates the use of more chemicals to cure; also, these chemicals are hazardous to other animals, insects, and birds that are beneficial to agriculture. Automatic detection of plant illnesses is critical for detecting disease symptoms in their early stages, when they occur on a plant's growing leaf.
Deep learning algorithms have been popular in recent years for picture categorization issues. Thanks to developments in artificial intelligence research, it is now possible to automatically diagnose plant diseases from raw photos. A learning strategy based on neural networks is known as deep learning[2].
This learning has the advantage of being able to automatically extract characteristics from photos. The neural network learns how to extract features during training. Therefore, by reducing the biotic factors that lead to significant agricultural yield losses, we can increase the productivity and quality of plants. Different machine learning and deep learning approaches can be used for this.

\section{Related Works}
In this section the previous research works related to image segmentation and classification in the domain of crop disease detection are discussed. To
segment coloured pictures genetic algorithm is used. The approach represents an image as a weighted undirected
network with edges signifying similar pixels and nodes denoting pixels. Brightness, colour, and texture richness are all considered when comparing the similarity of two pixels. Implementing effective procedures for identifying healthy and sick leaves aids in crop loss control and productivity. This section contains a collection of existing machine-learning approaches for identifying plant diseases.

\subsection{Shape and Texture-Based Identification}
The authors of \cite{article2} used tomato-leaf photos to identify illnesses. They classified sick segments using several geometric and histogram-based characteristics and an SVM classifier with varying kernels. S. Kaur et al.\cite{article5} discovered three distinct soybean diseases based on colour and textural characteristics.In \cite{thirtle2001relationship} P Babu et al. identified plant leaves and illnesses using a feed-forward neural network and backpropagation. S. S. Chouhan et al. \cite{inbook1} identified plant leaves and fungal infections using a bacterial-foraging-optimization-based radial-basis function neural network (BRBFNN). They employed a region-growing algorithm to extract information from a leaf based on seed points with similar qualities in their techniques. The bacterial-foraging optimization technique is used to increase classification accuracy and speed up a network.
\subsection{ Segmentation using Traditional Methods}
In article \cite{article3}, Vijai Singh introduced a system for photo segmentation that is used for automated diagnosis and classification of plant leaf diseases. The genetic technique utilised for picture segmentation, a crucial step in plant leaf disease identification. The average classification accuracy of the recommended algorithm, which was used to complete the classification, is 97.6
The authors of \cite{inproceedings1} proposed a three-step segmentation process. On green plants, the undifferentiated disease spots should be removed first. The split image is then transformed into 8-bit grayscale pictures using single thresholding after certain odd traits are found using a grey histogram. Finally, compare the size and stems of the sick spots, and then use area thresholding to segment the matching images.
\subsection{Deep-Learning-Based Identification}
Mohanty et al. Identified 26 distinct plant diseases using AlexNet and GoogleNet CNN architectures. Ferentinos et al.\cite{metcalfe1980crop} identified 58 distinct plant diseases using different CNN architectures, obtaining high levels of classification accuracy. They also tried the CNN architecture with real-time photos in their method. Sladojevic et al.\cite{article} created a deep learning framework to detect 13 distinct plant diseases. They trained CNN using the Caffe DL framework.. The authors of \cite{phdthesis} suggested a nine-layer CNN model to diagnose plant diseases. They employed the PlantVillage dataset and data-augmentation techniques to enhance the data size for experimentation purposes, and then examined performance.
\subsection{Leaf Image Classification}
According to a study in paper \cite{article}, segmenting and identifying diseases in live images are necessary for cardamom plant leaf disease identification. The recommended methodology's U2 -Net design achieves results without degrading the original image quality by removing the detailed backdrop. EfficientNetV2-S and EfficientNetV2-L achieved detection accuracy for the dataset of cardamom plants of 98.28\% and 98.26\%, respectively. In the study \cite{article1}, researchers proposed NAS-Unet, which is stacked by the equal quantity of DownSC and UpSC on a U-like backbone network. To accelerate the search, add a U-shaped backbone and the memory-saving Binary gate search method. Cell designs for semantic picture segmentation in this situation are DownSC and UpSC.

\section{Proposed Methodology}
There are many different methods for detecting plant diseases using machine learning, but most of them involve using a combination of image analysis and machine learning algorithms to identify the presence of diseases in plants. One common approach is to use convolutional neural networks (CNNs) to analyze images of plants and identify the presence of diseases. The CNNs are typically trained on a large dataset of labeled images, where the labels indicate whether a particular plant in the image is healthy or diseased.

\subsection{Dataset}\label{AA}
\begin{figure}[ht]
    \centering
    \centerline{\includegraphics[width=\linewidth]{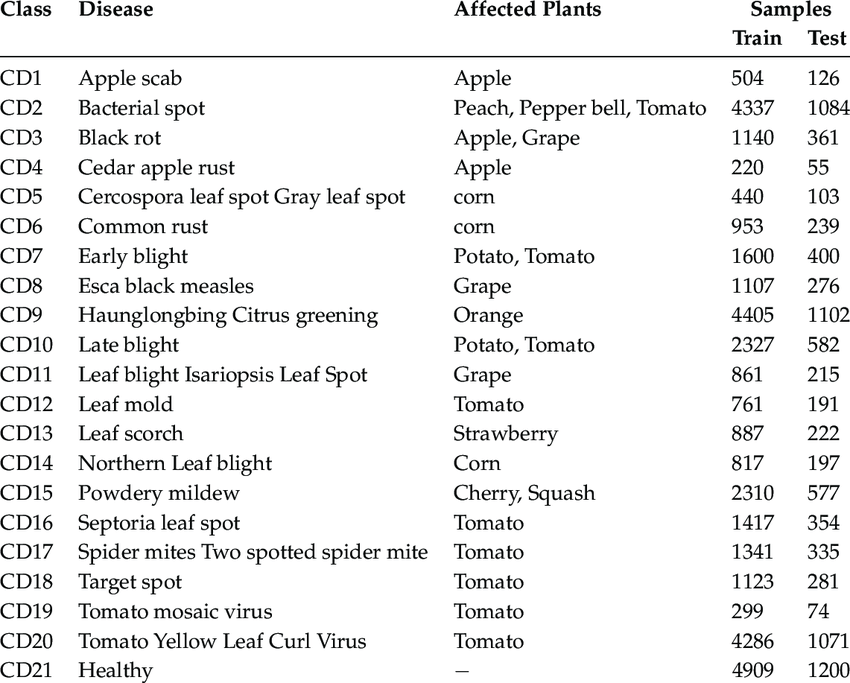}}
    \caption{Overview of PlantVillage Dataset}
    \label{fig:dataset}
\end{figure}

 In the PlantVillagedataset 38 plant crop-disease pairings, spanning 14 crop species and 26 diseases, can be found among the 54,305 leaf images as shown in Fig \ref{fig:dataset}. The clear photographs of plant leaves in this collection each feature a single leaf. Additionally, it provides pre-determined training and testing subsets, which we used in our study.
It was made using six distinct augmentation strategies to provide more varied datasets with various environmental factors. Scaling, rotation, noise injection, gamma correction, picture flipping, and PCA colour augmentation \ref{tab:Augmentation } were some of the augmentations employed in this procedure.
\subsection{Image Processing}  
The dataset needs to be resized because of the probability of an uneven size in images due to the dataset's size. To reduce noise and improve the image, gaussian blur is performed after downsizing the raw input photos to 256 x 256. Gaussian blur is a blurring effect commonly used in image processing and computer vision. As images are two dimensional, in 2-D form, a Gaussian kernel is represented as \ref{gauss}.The Gaussian blur filter works by convolving the original image with a Gaussian function. This function has a bell-shaped curve that is centered at the origin, which means that the filter has a uniform blurring effect across the entire image. 
\begin{equation}
    \label{gauss}
    G_{2 D}(a, b, \sigma)=\frac{1}{2 \pi \sigma^{2}} e^{-\frac{a^{2}+b^{2}}{2\sigma^{2}}}
\end{equation}
where a and b are the location indices,$\sigma$ is the distribution's standard deviation, and The Gaussian distribution's variance, which establishes the amount of the blurring effect surrounding a pixel, is controlled by the value of $\sigma$.\\
The amount of blurring can be controlled by adjusting the standard deviation of the Gaussian function, which determines how wide the bell-shaped curve is. In general, Gaussian blur is a useful tool for smoothing images and reducing noise. It is often used in computer vision algorithms to pre-process images before applying more complex operations. It is also commonly used in computational photography, where it can be used to create effects like depth of field and motion blur.
\begin{figure}[ht]
    \centering
    \centerline{\includegraphics[width = \linewidth ]{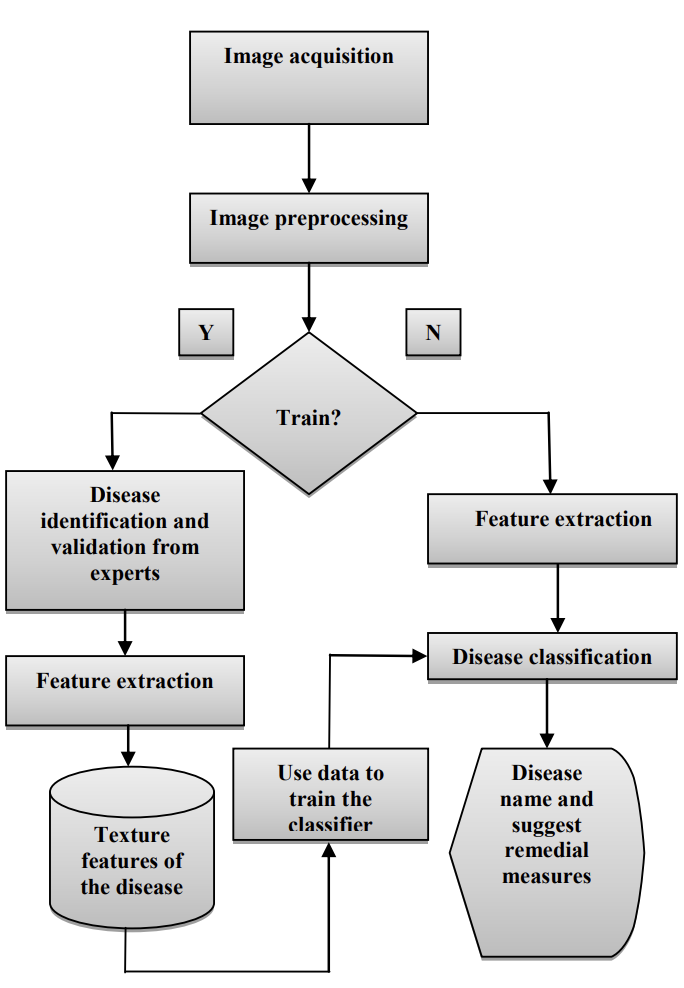}}
    \caption{Block diagram of proposed work}
    \label{fig:block}
\end{figure}

Through various processing techniques or combinations of multiple processing, such as random rotation, shifts, shear, and flips, etc., image augmentation artificially generates training pictures. The ImageDataGenerator API in Keras makes it simple to develop an enhanced picture generator. A real-time data augmentation tool called ImageDataGenerator creates batches of picture data.
\begin{table}[ht]
\caption{Data Augmentation Techniques}
\label{tab:Augmentation }
\begin{center}
\scalebox{0.9}

\begin{tabular}{|c|c|} 

\hline
\textbf{Technique} & \textbf{Hyperparameter}\\
\hline
 Scaling   & Yes\\
\hline
Rotation & Yes\\
\hline
Noise injection & Factor = 1\\
\hline
 Random Flip & Vertical \\
\hline
Gamma correction & Yes\\
\hline
PCA colour & 3 x 3\\
\hline
\end{tabular}
\end{center}
\end{table}
The following steps are included in the preparation of input images for the identification of healthy and diseased leaves using a neural network:
\begin{itemize}
    \item Image conversion to grayscale is frequently done to focus on the texture and form of the leaves rather than their colour and to reduce the number of dimensions in the input data. pixel values in a picture are normalised.
    \item Resize the images to a consistent scale (256 x 256) appropriate for the CNN model. This is frequently done to reduce the computational load of training the model and to ensure that all images have the same dimensions, which is essential for the CNN to properly analyze the images. 
    \item Image pixel values are normalized. This is frequently done to increase the performance of the CNN by ensuring that the pixel values are within a standard range (e.g., between 0 and 1) and have a zero mean and unit variance.
    \item Image enhancements are applied to the pictures. This entails applying random modifications to the images (e.g., rotating, scaling, etc.) in order to provide more training examples and increase the model's robustness.
\end{itemize}
The overview of the proposed methodology is shown in block diagram Fig \ref{fig:block}
\subsubsection{Image Thresholding}\label{sub:it}
Using image thresholding, a picture may be binarized depending on pixel intensities. Such a thresholding method typically takes as inputs a grayscale image and a threshold. What emerges are binary pictures. If the input pixel's intensity exceeds a threshold, the corresponding output pixel is labelled as white (foreground), and if it is equal to or less than the threshold, it is labelled as black (background). To find the spread for the pixel levels on each side of the threshold, or the pixels that are either in the foreground or background, Otsu's thresholding technique iterates over every possible threshold value. The objective is to determine the threshold value at which the sum of foreground and background spreads is at its minimum.
\begin{equation}
    \label{otsu}
    \sigma^2_w(t) = \omega_1(t)\sigma^2_1(t) +\omega_2(t)\sigma^2_2(t) 
\end{equation}

The whole computation equation of Otsu's thresholding can be defined by Eq \ref{otsu}, where weighted variance of classes denoted by $\sigma^2_w(t)$ .  $\omega_1(t)$ \text { and }$ \omega_2(t)$  are the probabilities of the two classes divided by a threshold $t$.\\
\subsubsection{Canny Edge Detector}\label{sub:ced}
Edge-based segmentation is a technique used in image processing to identify and extract the boundaries of objects in an image. This is typically accomplished by applying a filter to the image that emphasizes the edges and transitions between different regions of the image, such as the boundaries between foreground and background objects. The output of an edge-based segmentation algorithm is a set of edge maps, which are binary images that highlight the locations of edges in the input image. These edge maps can then be used as input to other image processing algorithms, such as object recognition and tracking algorithms, to identify and classify the objects in the image.
An edge detector with several stages is the Canny filter. A filter with a Gaussian derivative is used to determine the gradients' intensities. The Gaussian reduces the amount of noise in the picture. Edge pixels are either kept or discarded using hysteresis thresholding that is applied to the gradient magnitude.
The Canny contains three variables that may be altered: the Gaussian's width (the larger the Gaussian, the noisier the picture), as well as the low and high thresholds for hysteresis thresholding.
After preprocessing the input images, they may be fed into the CNN model for training and testing.
\subsection{Training}

The preprocessed are now splitted into train test splits in the ratio of 80:20 and 
\begin{figure}[ht]
    \centering
    \centerline{\includegraphics[width = \linewidth ]{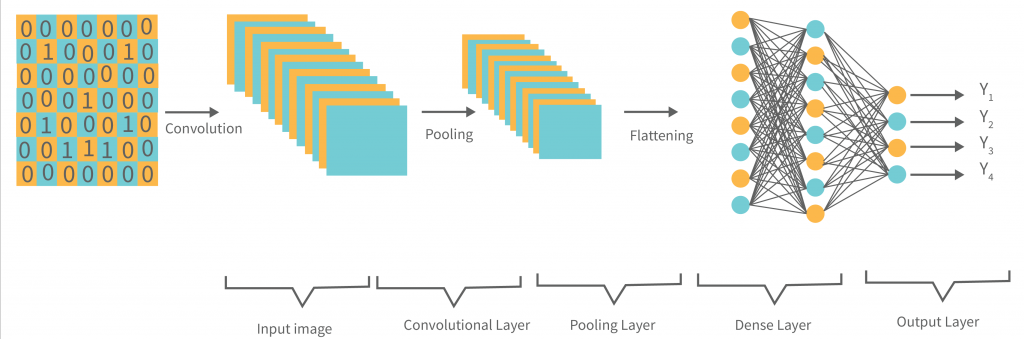}}
    \caption{Architecture of CNN}
    \label{fig:cnnar}
\end{figure}
after splitting the training is being started with some initial hyper parameters. CNNs have lately gained popularity, and DL is the most common architecture because DL models can learn significant characteristics from input images at different convolutional levels, similar to how the human brain works. With high classification accuracy and a low mistake rate, DL can solve complicated problems exceptionally successfully and rapidly \cite{article4}. The main components of a CNN are the convolutional layers, pooling layers, and fully connected layers. The convolutional layers apply a set of filters to the input data, each of which is designed to detect a specific feature or pattern in the data. 
\begin{figure}[ht]
    \centering
    \centerline{\includegraphics[width = \linewidth ]{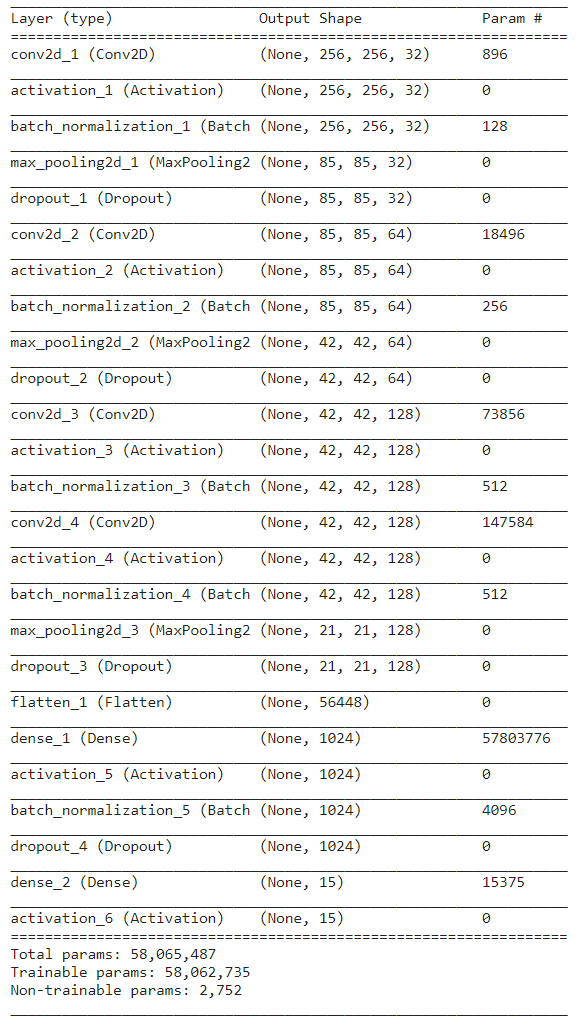}}
    \caption{Summary of the trained CNN Model}
    \label{fig:modelsumm}
\end{figure}
The main components of a CNN are the convolutional layers, pooling layers, and fully connected layers. The pooling layers reduce the dimensionality of the data by downsampling the output of the convolutional layers. The fully connected layers combine the features detected by the convolutional and pooling layers and use them to make predictions about the input data. The key advantage of CNNs is that they can learn to automatically detect and extract features from the input data, which reduces the need for manual feature engineering. This makes them well-suited for tasks such as image classification, object detection, and segmentation.

The summary of the CNN model built for classification of images is shown in Fig \ref{fig:modelsumm}.
CNN is a type of deep learning neural network that is designed to process data with a grid-like structure.

 It consists of multiple layers of interconnected nodes, each of which performs a specific mathematical operation on the input data. The Architecture the CNN is shown in Fig \ref{fig:cnnar}. It's important to note that training a neural network for a certain number of epochs is just one aspect of the overall training process. In order to achieve good performance, it's also necessary to carefully select and preprocess the training data, and to tune the network's hyperparameters to optimize its performance. Additionally, the network's performance may improve if it is trained for more epochs, although this will also increase the time and computational resources required for training.
\section{Results}
Leaf disease classification involves identifying the type of disease that is affecting a plant's leaves. This is typically done by visually inspecting the leaves and observing the symptoms they exhibit, such as discoloration, spotting, or wilting. The specific symptoms can then be used to determine the type of disease that is present.
The result of applying the Canny edge detection algorithm to an image is a binary image where pixels that correspond to edges in the original image are marked as "on" (usually represented as white pixels), and all other pixels are marked as "off" (usually represented as black pixels). 
\begin{figure}[ht]
    \centering
    \centerline{\includegraphics[width = \linewidth ]{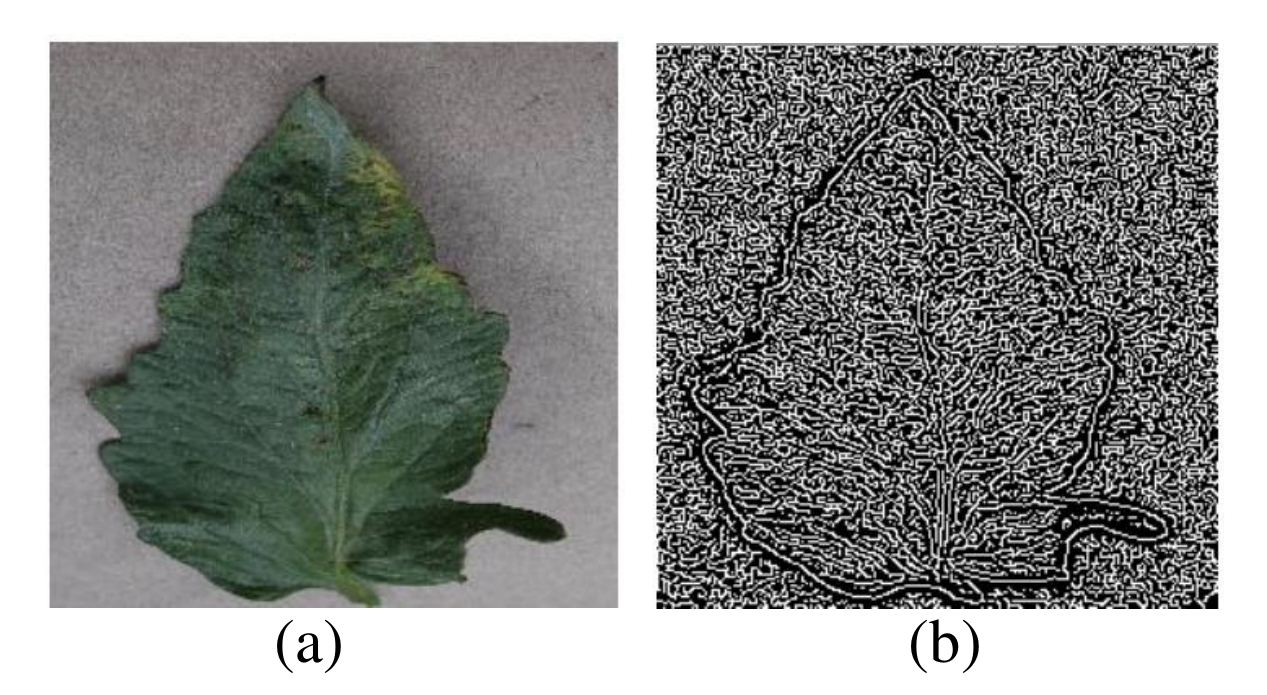}}
    \caption{(a) Input image (b) result of canny edge detection}
    \label{fig:canny}
\end{figure}
The resulting image can then be used for further image processing tasks, such as object recognition or image segmentation. The result of canny edge algorithm is shown in Fig \ref{fig:canny}

\begin{figure}[ht]
    \centering
    \centerline{\includegraphics[width = \linewidth ]{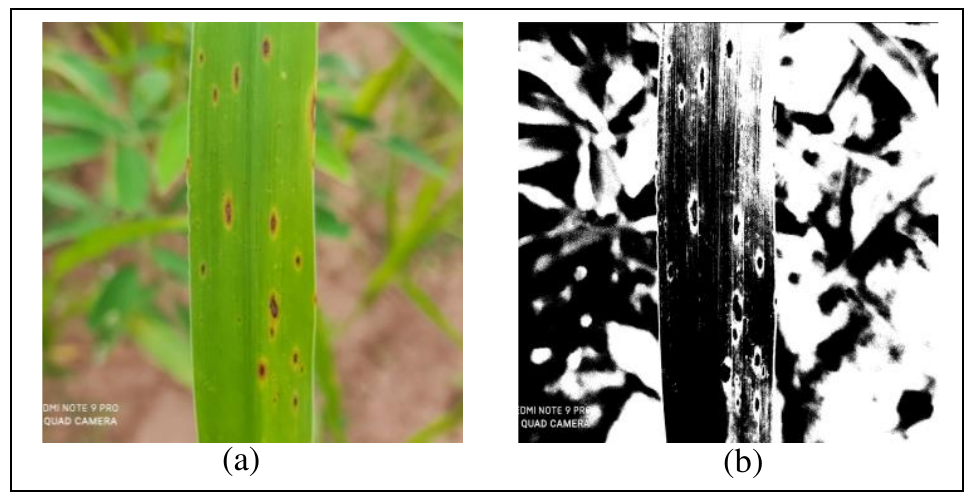}}
    \caption{(a) Input image (b) result of image thresholding}
    \label{fig:ostuss}
\end{figure}
The result of applying image thresholding is a binary image that preserves the important features of the original image while eliminating most of the noise and reducing the amount of data that needs to be processed. The result of image thresholding algorithm is shown in Fig \ref{fig:canny}

This can be useful for a variety of image processing tasks, such as object recognition or image segmentation.
After 30 epochs of training, a CNN will have learned to recognize and classify objects in images with a high degree of accuracy. The exact performance of the network will depend on several factors, including the quality and diversity of the training data, the specific architecture of the network, and the hyperparameters used during training. In general, though, a well-trained CNN should be able to achieve high accuracy on a wide range of image recognition tasks after 30 epochs of training.
\begin{figure}[ht]
    \centering
    \centerline{\includegraphics[width = \linewidth ]{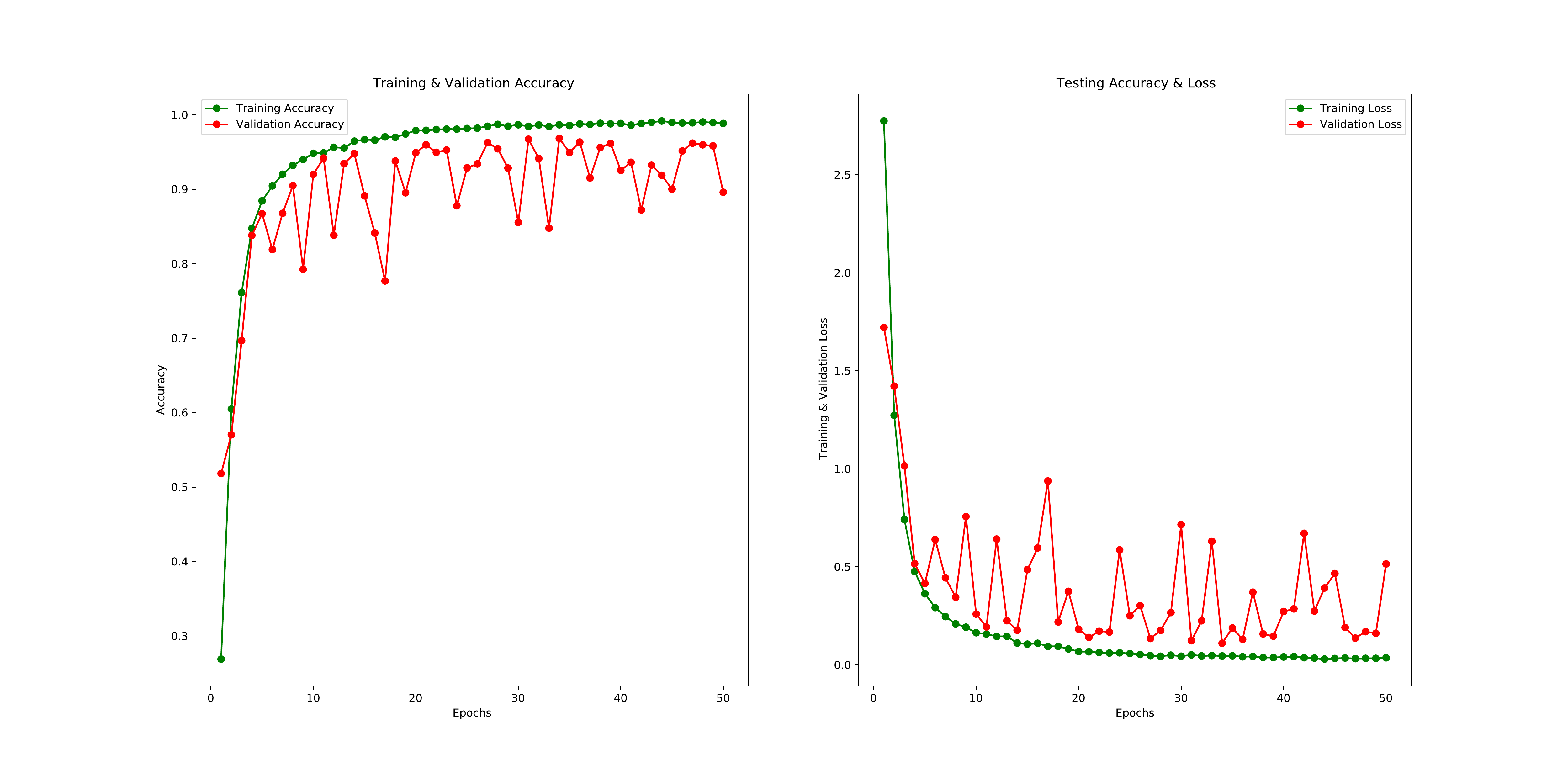}}
    \caption{Graph depecting training and validation accuracy and loss}
    \label{fig:graphs}
\end{figure}
 The training results are shown in Fig \ref{fig:graphs}
It's important to note that training a neural network for a certain number of epochs is just one aspect of the overall training process. In order to achieve good performance, it's also necessary to carefully select and preprocess the training data, and to tune the network's hyperparameters to optimize its performance. Additionally, the network's performance may improve if it is trained for more epochs, although this will also increase the time and computational resources required for training.
A confusion matrix is a table that is used to evaluate the performance of a classification model, typically in the context of supervised machine learning. 
\begin{figure}[ht]
    \centering
    \centerline{\includegraphics[width = \linewidth ]{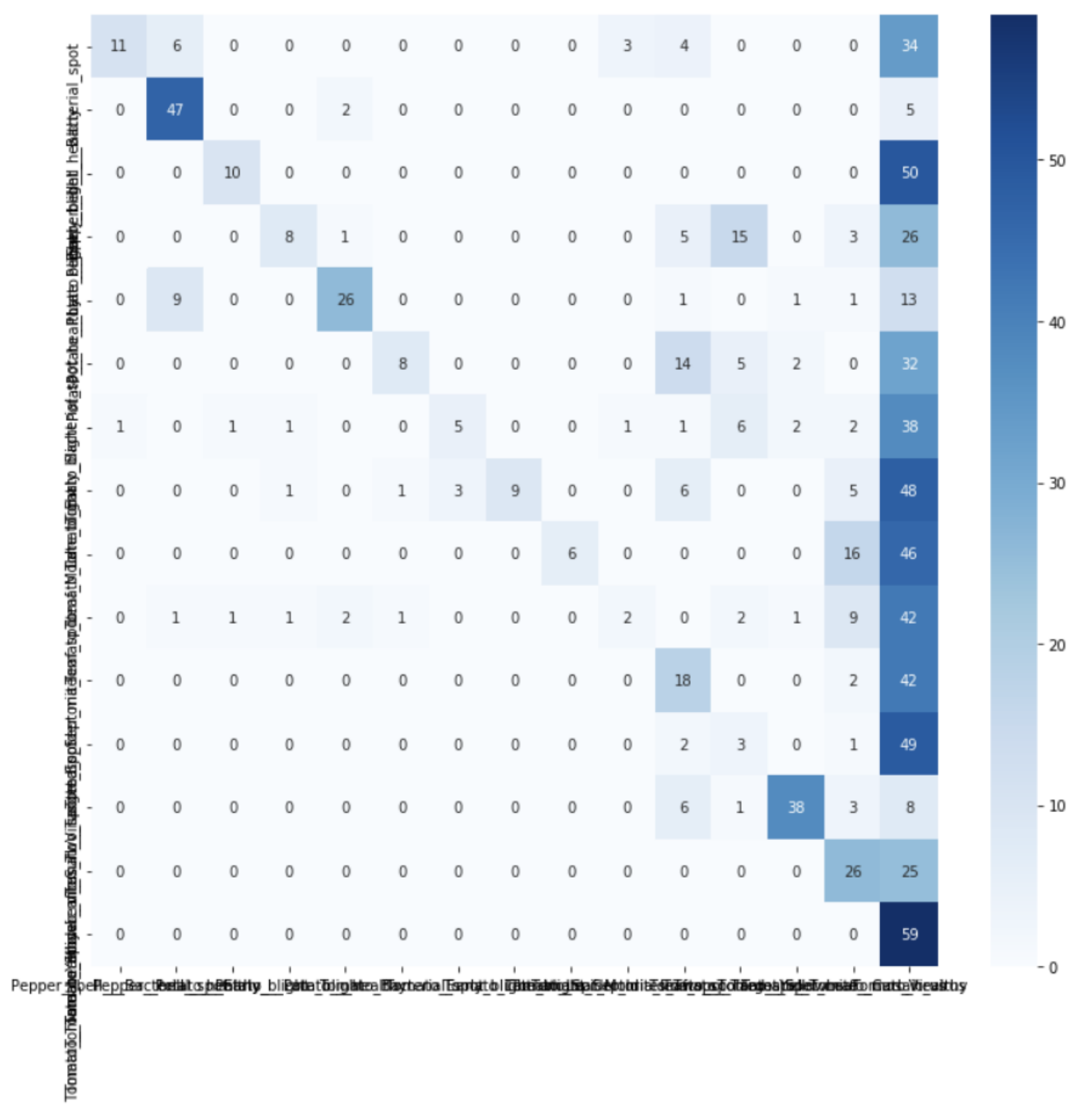}}
    \caption{(Confusion matrix of predicted classes}
    \label{fig:conf}
\end{figure}
The matrix is used to evaluate the model's ability to correctly classify data into one of several classes, and is often used in conjunction with classification metrics such as accuracy, precision, and recall. The confusion matrix of above training is shown in Fig \ref{fig:conf}
A confusion matrix for a classification model with 15 classes would have 15 rows and 15 columns, with each row representing the predicted class and each column representing the actual class. The entries in the matrix would show the number of instances in the test dataset that were predicted to belong to a particular class and actually belong to that class, as well as the number of instances that were predicted to belong to a particular class but actually belong to another class.
\section{Conclusion}
We conclude from this work that disease detection and classification can be done using image processing. The suggested approach can classify  crop diseases extremely precisely and effectively. The suggested approach was designed with farmers and the agricultural sector in mind. The created technology will detect plant disease and provide corrective action. Proper knowledge of the disease and the therapy can be used to improve the plant's health. It is possible to use digital image processing techniques to perform image classification of diseased leaves of 15 classes. By training a suitable machine learning model on a dataset of labeled leaf images, it is possible to develop a system that can automatically identify and classify the different types of diseases present in the leaves.
The performance of the classification model will depend on several factors, including the quality and diversity of the training data, the specific machine learning algorithm used, and the hyperparameters chosen for the model. In general, though, a well-trained model should be able to achieve high accuracy on a variety of image classification tasks. Using digital image processing techniques to classify diseased leaves can provide a number of benefits, including improved diagnostic accuracy, faster and more efficient disease detection, and the ability to automate the classification process. This can help to reduce the time and resources required to diagnose and treat plant diseases, and can ultimately improve the overall health and productivity of agricultural crops.

\section*{Acknowledgment}
This research was supported by International Institute of Information Technology, Naya Raipur. we would like to thank Dr. Kanad Biswas , Bennett University (formerly with IIT Delhi) and Abhishek Sharma , assistant professor at International Institute of Information Technology, Naya Raipur who
guided and gave us support through out this research.

\bibliography{references.bib}

\end{document}